\documentclass[conference]{IEEEtran}
\IEEEoverridecommandlockouts
\usepackage{cite}
\usepackage{amsmath,amssymb, amsbsy,bm,bbm,amsthm}
\usepackage{algorithm, algorithmic}
\usepackage{graphicx}
\usepackage{textcomp}
\usepackage{xcolor}
\usepackage{adjustbox,multirow, booktabs, hhline,tabularx, multirow}
\usepackage{subcaption, xcolor}
\usepackage{hyperref}

\def\1{\mathbf 1}

\def \bdelta{\bm{\delta}}
\DeclareMathOperator*{\argmin}{\arg\!\min}
\DeclareMathOperator*{\argmax}{\arg\!\max}

\def\BibTeX{{\rm B\kern-.05em{\sc i\kern-.025em b}\kern-.08em
    T\kern-.1667em\lower.7ex\hbox{E}\kern-.125emX}}
\begin{document}

\title{MGSER-SAM: Memory-Guided Soft Experience Replay with Sharpness-Aware Optimization for Enhanced Continual Learning
}

\author{\IEEEauthorblockN{1\textsuperscript{st} Xingyu Li}
\IEEEauthorblockA{\textit{Department of Computer Science} \\
\textit{Tulane University}\\
New Orleans, United States \\
xli82@tulane.edu}
\and
\IEEEauthorblockN{2\textsuperscript{nd} Bo Tang}
\IEEEauthorblockA{\textit{Department of Electrical and Computer Engineering} \\
\textit{Worcester Polytechnic Institute}\\
Worcester, United States \\
btang1@wpi.edu}
}

\maketitle

\begin{abstract}
    Deep neural networks suffer from the catastrophic forgetting problem in the field of continual learning (CL). To address this challenge, we propose MGSER-SAM, a novel memory replay-based algorithm specifically engineered to enhance the generalization capabilities of CL models. We first intergrate the SAM optimizer, a component designed for optimizing flatness, which seamlessly fits into well-known Experience Replay frameworks such as ER and DER++. Then, MGSER-SAM distinctively addresses the complex challenge of reconciling conflicts in weight perturbation directions between ongoing tasks and previously stored memories, which is underexplored in the SAM optimizer. This is effectively accomplished by the strategic integration of soft logits and the alignment of memory gradient directions, where the regularization terms facilitate the concurrent minimization of various training loss terms integral to the CL process. Through rigorous experimental analysis conducted across multiple benchmarks, MGSER-SAM has demonstrated a consistent ability to outperform existing baselines in all three CL scenarios. Comparing to the representative memory replay-based baselines ER and DER++, MGSER-SAM not only improves the testing accuracy by $24.4\%$ and $17.6\%$ respectively, but also achieves the lowest forgetting on each benchmark.
\end{abstract}


\section{Introduction}\label{sec:intro}
Recently, Deep Learning has shown great successes in solving human-like tasks, such as image classification \cite{krizhevsky2012imagenet}, security \cite{li2021lomar}, and distributed optimization \cite{li2023fedlga}. However, when the model is trained sequentially with new tasks drawn from a continuous or streaming scheme, a.k.a., continual learning (OCL), the performance decreases dramatically due to the loss of learned information, called the \textit{catastrophic forgetting} \cite{goodfellow2013empirical}. To address this problem, there have been an abundant number of approaches, which can be typically divided into three types: regularization \cite{kirkpatrick2017overcoming, zenke2017continual}, architectural \cite{rusu2016progressive}, and replay \cite{lopez2017gradient, rolnick2018experience}. 

Though there have been debates about the extra storage of seen experiences, e.g., training data points in the replay-based methods, the efficiency of which has been well-accepted, especially within the challenging contexts of the three standard CL scenarios, task-incremental (task-IL), domain-incremental (domain-IL), and class-incremental (class-IL) \cite{van2019three}. Typically, replay-based methods, such as Experience Replay (ER), involve two main stages: update and replay, where the seen data points is stored into a memory buffer via reservoir sampling \cite{vitter1985random} and randomly selected for replay when learning the new tasks through stochastic gradient optimization (SGD).  

Typically, the objective function of replay-based CL method can be regarded as the combination of the current task loss and the memory buffer loss, e.g., cross-entropy on the image classification tasks. Conventional SGD optimization minimizes the losses empirically, where the model parameters are updated in the direction of the averaged training and memory data points, which can fall into suboptimal minima and leads to overfitting. In this paper, we aim to enhance the replay-based CL methods, startign from the first milestone, ER, which is the most representative replay-based CL method. Our motivation is driven by the recent studies \cite{keskar2017on,chaudhari2019entropy,qu2022generalized} with the claim that the model performance can be improved by improving the model generalization capability, which can be achieved by flatting the geometry of the loss function.

	\begin{figure}[t!]
		\centering
		\includegraphics[width=0.99\columnwidth]{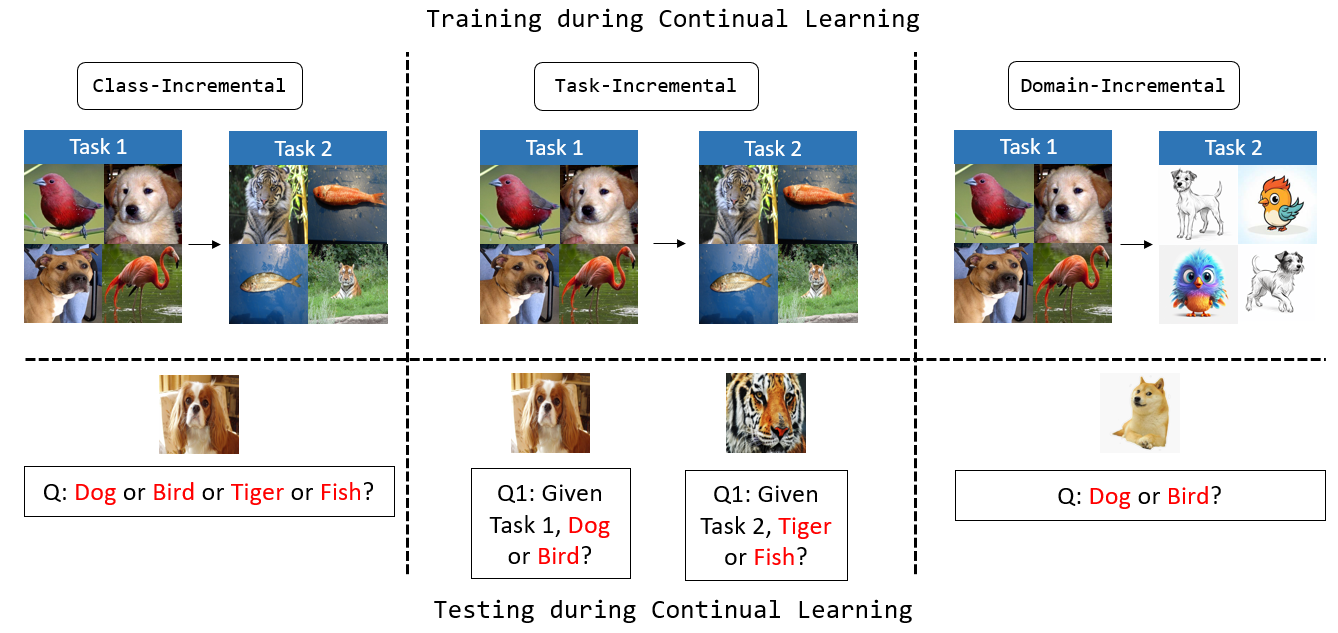}
		\caption{Continual learning three scenarios example.}
		\label{fig:scenario}
	\end{figure}

To achieves this, we explore the cutting-edge technique of sharpness-aware minimization (SAM) \cite{foret2020sharpness}, a recent development in the optimization landscape which is crafted to enhance model flatness by minimizing the worst-case loss in the immediate vicinity of the model parameter space, achieved through weight perturbation which remarkably requires only linear extra computational complexity. Our first introduce the ER-SAM algorithm, where the SAM optimizer is seamlessly integrated with ER, functioning as a component for optimizing flatness. Note that our integration is designed for scalability, making it applicable to existing replay-based methods.

Upon further investigation of the challenges arising from the combined optimization objective inherent in ER, we have proposed a novel algorithm named Memory Guided Soft Experience Replay with Sharpness-Aware Minimization (MGSER-SAM). This algorithm adeptly addresses the potential conflicts between the weight perturbation directions associated with the current task and those in the memory buffer, a unique challenge stemming from the non-stationary nature of continual learning. MGSER-SAM ingeniously incorporates soft logits and aligns memory gradient directions, providing effective regularization terms to enable the simultaneous minimization of various training loss terms, encompassing both current tasks and memory integration within the CL process. Through rigorous experimental analysis across a spectrum of benchmarks, the superiority of MGSER-SAM has been demonstrated over existing baselines in all three CL scenarios. Specifically, when compared with representative baselines such as ER and DER++ \cite{rolnick2018experience,buzzega2020dark}, MGSER-SAM not only significantly enhances testing accuracy by $24.4\%$ and $17.6\%$ respectively, but also achieves the lowest levels of forgetting. 


The main contributions of this paper are as follows: (1) We introduce the ER-SAM algorithm, which effectively applies the SAM optimizer to ER. The SAM optimizer is scalable to be integrated with a wide range of ER-like methods. (2) We proposed the novel MGSER-SAM algorithm, adept at resolving conflicts between weight perturbation directions in current tasks and memory buffer. (3) The conducted experimental results across multiple benchmarks show the scalability of the provided SAM optimizer to replay-based CL methods, and the superiority of the proposed MGSER-SAM algorithm across various benchmarks in all three CL scenarios.

\section{Related Work}\label{Sec:related}

\subsection{Three Scenarios of Continual Learning}
Evaluating continual learning approaches poses notable challenges, stemming from variations in experimental protocols and the degree of access to task identity during testing. In response to these challenges, \cite{van2019three} made a significant contribution by introducing three standardized evaluation scenarios, which has gained widespread adoption in the research community. This is evidenced by their utilization in subsequent studies such as \cite{buzzega2020dark}.

In this paper, we present an image classification benchmark demonstration to show the three scenarios of continual learning. Our model is sequentially trained on a variety of animal images, including birds, dogs, tigers, and fish, as illustrated in Figure~\ref{fig:scenario}. The first scenario, task-incremental learning (\textbf{task-IL}), challenges the model to correctly identify classes within a specified task. A more complex scenario is the class-incremental learning (\textbf{class-IL}), wherein the learner must discriminate between all previously encountered animal classes. Diverging slightly, the domain-incremental learning (\textbf{domain-IL}) scenario, as discussed in \cite{parisi2019continual,van2019three}, involves tasks that contain classes with identical labels but exhibited in varying domains. For instance, task 1 may present real-world images of birds and dogs, while task 2 showcases these animals in a cartoon style, as depicted in Figure~\ref{fig:scenario}. The primary objective of this paper is to enhance the performance of continual learning models by thoroughly considering and addressing these three scenarios.

\subsection{Three Types of Continual Learning Approache}
\subsubsection{Regularization}
Existing continual learning (CL) approaches can also be divided into three categories based on the model training process: regularization, architectural, and replay. 
Regularization methods are designed to alleviate the issue of forgetting by striking a balance between retaining knowledge from previous tasks and assimilating new information from the current task. Learning without Forgetting (LwF) \cite{li2017learning}, a notable technique in this domain, accomplishes this through a knowledge distillation process where insights from a larger model is transferred to a smaller one, ensuring that predictions for the current task are aligned with previously learned tasks. Elastic Weight Consolidation (EWC) \cite{kirkpatrick2017overcoming} takes a different approach by using Fisher Information to pinpoint critical weights for past examples, subsequently applying constraints on their alterations. Synaptic Intelligence (SI) \cite{zenke2017continual} introduces a unique method of penalizing parameters within the model objective function, differentially based on gradient information to identify key parameters. ISYANA \cite{mao2021continual} extends this concept by examining both the inter-task relationships and the connections between various concepts within the model. A recent innovation \cite{sun2023exemplar} utilizes variational auto-encoders to enable exemplar-free continual learning.


\subsubsection{Architectural}
Dynamic architectural approaches in CL adapt the structure of the model itself to accommodate new tasks, primarily through the addition of new neural resources, e.g., layers. \cite{yao2010boosting} implements a linear increase in the number of models corresponding to the introduction of new tasks. Progressive Neural Networks (PNN) \cite{rusu2016progressive} takes a different approach by preserving the architecture of the previously trained network and augmenting it with new sub-networks with a fixed capacity dedicated to learning new information. Another innovative method is the Dynamically Expanding Network (DEN) \cite{yoon2017lifelong}, which incrementally enlarges the number of trainable parameters to better adapt to new examples, thus offering an online solution for increasing network capacity. 

\subsubsection{Memory Replay}
Replay-based methods in CL are fundamentally inspired by the relationship observed in neuroscience between the mammalian hippocampus and neocortex. 
\cite{shin2017continual} introduced a dual-model architecture, which consists of a generative component alongside a continual learning solver, facilitating the sampling and interleaving of trained examples through a process termed Generative Replay (GR). Gradient Episodic Memory (GEM) \cite{lopez2017gradient} innovatively stores a subset of encountered examples as episodic memory, which positively influences previous tasks. Its derivative, Averaged GEM (A-GEM) \cite{chaudhry2018efficient}, enhances the computational and memory efficiency of GEM using an averaging mechanism. Experience Replay (ER) \cite{rolnick2018experience} employs reservoir sampling \cite{vitter1985random} to maintain the memory buffer, thus approximating the data distribution. Further advancements in this field, such as \cite{aljundi2019online}, have refined the memory update and replay processes for improved performance. Meanwhile, FoCL \cite{lao2021focl} diverges by focusing on regularizing the feature space instead of the parameter space. \cite{li2021lomar} havs addressed the instability issues in ER with advanced memory buffer selection and update strategies.

Moreover, replay-based continual learning methods have been integrated with other machine learning paradigms like Meta-learning \cite{martins2023meta}, further broadening their applicability. Despite their memory-intensive nature, replay-based methods have consistently demonstrated high performance. In this paper, our focus is on augmenting the effectiveness of replay-based methods from two angles: enhancing the model's generalization capability and refining the rehearsal of distinct features within the memory space.

\section{Preliminaries}
\label{sec:preliminary}

Typically, the goal of a deep neural network in conventional deep learning if to train a model $f$ parameterized by $\bm{\theta}$ to an optimal state $\bm{\theta}^{*}$, targeting machine learning problems such as image classification over training dataset $\mathcal{X} = \{\bm{x}, y\}$ which is sampled from a statistical distribution $\mathcal{D}$ that follows:
\begin{equation}\label{eq:stationary}
	\bm{\theta}^{*}  = \argmin_{\bm{\theta}} \mathbb{E}_{(\bm{x}, y) \sim \mathcal{D}} [l(f_{\bm{\theta}}(\mathbf{x}, y)],
\end{equation}
where $l(\cdot, \cdot)$ is the loss function, e.g., cross-entropy for the empirical risk of $f(\bm{\theta})$ over $\mathcal{D}$. However, in continual learning, this stationary optimization process can be challenging, as the learning process is split into multiple tasks in a sequential and non-static manner. We define a CL problem as a sequence of $T$ tasks, indexed as $\mathcal{X}_t \sim \mathcal{D}_t, t \in [1, \dots, T]$. For the $t$-th task, the model  encounters training data points exclusively from  $\mathcal{X}_t$. While the conventional learning process methodologies efficiently identify the optimal $\bm{\theta}^{*}$ for task t, they overlook retaining previously acquired knowledge. This phenomenon leads to a significant challenge known as catastrophic forgetting \cite{goodfellow2013empirical}, where the model's ability to perform on prior classification tasks diminishes as it adapts to new information. We denote the distribution of previously learned tasks as $\mathcal{X}_s = \{\mathcal{X}_1, \dots, \mathcal{X}_{t-1}\}$, then the optimal state of a CL model to simultaneously learn the new seen task and mitigate the forgetting issue can be formulated as follows:
\begin{equation}\label{Eq:continual}
	\theta^{*}  = \argmin_{\theta} (\mathcal{L}_{t} + \mathcal{L}_{s}),
\end{equation}
where $\mathcal{L}_{t}$ is the loss function for the current t-th task $\mathbb{E}_{(\mathbf{x}_t, y_t)\sim \mathcal{D}_t} [l(f_\theta(\mathbf{x}_t), y_t)]$, and $\mathcal{L}_{s}$ is the loss function for the learned tasks. Note that the challenge of simultaneously minimizing $\mathcal{L}_{t}$ and $\mathcal{L}_{s}$ is also known as the \textit{stability-plasticity dilemma} \cite{lopez2017gradient}, where stability requires good performance on $\mathcal{X}_s$ and plasticity is the quick adaptation to $\mathcal{X}_t$. 

To achieve the objective in Eq.~\eqref{Eq:continual}, \cite{rolnick2018experience} provides a memory-based approach called Experience Replay, where a $M$-sized memory buffer $\mathcal{M}= \{(\mathbf{x}_1, y_1), \dots, (\mathbf{x}_m, y_m), \dots (\mathbf{x}_M, y_M)\}$. Then, when training the $t$-th task, ER samples a mini-batch of data points from $\mathcal{M}$ as $\mathcal{B}$ and merges it with the current task training batch, which address the optimization objective \eqref{Eq:continual} with the following loss function:
\begin{equation}\label{Eq:er}
	\mathcal{L}_{total} = \mathbb{E}_{(\mathbf{x}, y) \sim \mathcal{D}_t} [l(f_\theta(\mathbf{x}), y)]+ \mathbb{E}_{(\mathbf{x}, y) \sim \mathcal{B}} [l(f_\theta(\mathbf{x}), y)].
\end{equation}
Note that as $\mathcal{M}$ is updated during the learning of sequential tasks with data coming like a stream, ER adopts reservoir sampling \cite{vitter1985random} to provide a guarantee that each data point in $\mathcal{D}_s$ has the same probability to be sampled into $\mathcal{M}$. ER has been considered as a practical solution for CL, as it is easy to implement and has a good performance. However, the performance of ER can be further improved regarding the memory buffer quality and the continual learning algorithm design. Specifically, Eq.~\eqref{Eq:er} simply minimizes the empirical loss, 
which can lead to overfitting that limits the generalization capability of the model. Recent studies \cite{keskar2017on,chaudhari2019entropy,qu2022generalized} have shown that flatting the geometry of the loss function can improve the model generalization ability to mitigate the overfitting problem. Hence, in this paper, we aim to improve the performance of experience replay by leveraging the geometry of the loss function for better generalization capability.

\section{Method}
\label{sec:method}
In this section, we propose Memory Guided Soft Experience Replay with Sharpness-Aware Minimization (MGSER-SAM) algorithm. We will first present the direct application of SAM on ER, called ER-SAM, which can be considered as a flatness component that is orthogonal and can be added to existing ER like methods for generalization with flat minima. Next, we show our conjecture about the potential conflict between the weight perturbation directions from current task and memory. To this end, we present our proposed MGSER-SAM algorithm, which further improves ER by leveraging the soft logits and memory gradient directions  as regularization terms to guide the sharpness-aware minimization update direction.

\subsection{Experience Replay with Sharpness Aware Minimization (ER-SAM)}
As illustrated in Sec.~\ref{sec:preliminary}, our goal is to improve the generalization capability of ER whose loss function consists of two parts: the current task loss and the memory loss. Sharpness-aware minimization \cite{foret2020sharpness} is a recently developed technique that minimizes the worst-case loss in a neighborhood of the current parameter $\bm{\theta}$, which can be a natural solution for the generalization of experience replay. Given the model $\bm{\theta}$, the SAM loss function in ER can be formulated as follows:
\begin{equation}\label{Eq:ersam}
	\min_{\bm{\theta}} \max_{\|\bdelta \|_2 \leq \rho} L_{total}(\bm{\theta}+\bdelta),
\end{equation}
where $\rho$ is a predefined constant that controls the radius of the neighborhood, and $\bdelta$ is the weight perturbation that maximizes the worst-case loss of $\mathcal{L}_{total}$ within the Euclidean ball around $\bm{\theta}$. The optimal $\bdelta^{\star}$ can be found by turning into the following linear constrained optimization with the first-order Taylor expansion of $L_{total}(\bm{\theta}+\bdelta)$ around $\bm{\theta}$ via one back-propagation operation, which can be obtained via one extra step of back-propagation operation that
\begin{equation}\label{Eq:sam}
	\begin{split}
	   \bdelta^{\star} & \triangleq \argmax_{\|\bdelta \|_2 \leq \rho} \mathcal{L}_{total} (\bm{\theta} + \bdelta ) \\
		& \approx \argmax_{\|\bdelta \|_2 \leq \rho} \mathcal{L}_{total} (\bm{\theta}) + \bdelta^{\top} \nabla_{\bm{\theta}} \mathcal{L}_{total} (\bm{\theta})  + O(\rho^2 ) \\
		& \approx \rho \frac{\nabla_{\bm{\theta}} \mathcal{L}_{total} (\bm{\theta})}{\|\nabla_{\bm{\theta}} \mathcal{L}_{total} (\bm{\theta})\|_2},
	\end{split}
\end{equation}	
where $\mathcal{O}(\rho^2)$ denotes the remainder of higher terms in the Taylor expansion of $\mathcal{L}_{total}$, which is ignored and not discussed in this paper, as it has limited impact to the approximation. In this case, the ER-SAM model update can be written as:
\begin{equation}\label{Eq:ersam-update}
	\begin{split}
	& \bm{g}^{ER-SAM}  = \nabla_{\bm{\theta}} \mathcal{L}_{total} (\bm{\theta}) |_{\bm{\theta}+ \bdelta^{\star}}, \\
	& \bm{\theta}_{t+1}  = \bm{\theta}_{t} - \eta \bm{g}^{ER-SAM},  \\
	\end{split}
\end{equation}
where $\eta$ is the learning step. To present the difference between \texttt{ER} and \texttt{ER-SAM}, we summarize the training procedures in Algorithm~\ref{ALG:ersam}. The SAM optimizer shares a similar idea of adversarial training with diverged focus. Rather than perturbing data points, it concentrates on weight perturbation, specifically targeting the worst-case loss neighborhood.

\begin{algorithm}[t!]
	\caption{\colorbox[rgb]{1.0, 0.55, 0.41}{ER} and \colorbox[rgb]{0.74,0.83,1}{ER-SAM}}
	\begin{algorithmic}
	 \STATE Initialized $f_{\bm{\theta}}$, Memory buffer $\mathcal{M}$, Learning rate $\eta$.
	 \FOR {tasks $t = 1, \dots, T$}
	 \STATE Sample batch $\mathcal{B}_t$ for task $t$ from  $\mathcal{D}_t$.
	 \STATE Sample batch $\mathcal{B}$ from memory buffer $\mathcal{M}$.
	 \STATE Merge training batches $\mathcal{B} = \mathcal{B}+\mathcal{B}_t$.
	 \STATE Compute the gradient $\bm{g} = \nabla_{\bm{\theta}} \mathcal{L}_{total} (\bm{\theta})$ with $\mathcal{B}$.
	 \STATE \colorbox[rgb]{1.0, 0.55, 0.41}{$\bm{\theta}_{t+1} = \bm{\theta}_{t} - \eta \bm{g}$.} 
	 \STATE \colorbox[rgb]{0.74,0.83,1}{Compute local model $\bm{\theta}_{t+1}$ from \eqref{Eq:ersam-update}.}
	 \ENDFOR
	\end{algorithmic}
	\label{ALG:ersam}
 \end{algorithm}

\subsection{Memory Guided Soft Experience Replay with Sharpness-Aware Minimization: MGSER-SAM}
However, different from SAM in conventional machine learning that flatting the stationary loss function, the $\mathcal{L}_{total}$ in this paper is a combination of the current task loss $\mathcal{L}_t$ and the memory loss $\mathcal{L}_{s}$. The previous developed ER-SAM simply minimizes the $\mathcal{L}_{total}$, which overlooks the potential conflict between $\mathcal{L}_{s}$ and $\mathcal{L}_t$. Inspired by \cite{tran2023sharpness},  we re-formulate $\bm{g}^{ER-SAM}$ with the consideration of both loss terms as: 
\begin{equation}\label{Eq:gsam_formulate}
	\begin{split}
	 \bm{g} & = \nabla_{\bm{\theta}} \mathcal{L}_{total} (\bm{\theta}+ \bdelta^{\star}), \\
	& \overset{(a)}{\approx} \nabla_{\bm{\theta}}[\mathcal{L}_{total} (\bm{\theta}) + \langle \bdelta^{\star},\nabla_{\bm{\theta}} \mathcal{L}_{total}(\bm{\theta})  \rangle ] \\
	& = \nabla_{\bm{\theta}}[\mathcal{L}_{total} (\bm{\theta})  +  \rho \|\nabla_{\bm{\theta}} \mathcal{L}_{total}(\bm{\theta})\|_2   ]\\
	& \leq \nabla_{\bm{\theta}}[\mathcal{L}_{t} (\bm{\theta}) +\mathcal{L}_{s} (\bm{\theta}) + \rho ( \|\nabla_{\bm{\theta}} (\mathcal{L}_{t}+\mathcal{L}_{s})(\bm{\theta})\|_2], \\ 
	\end{split}
\end{equation}
where $\bm{g}$ is the short for $\bm{g}^{ER-SAM}$, and the step in (a) is the first-order Taylor expansion. This observation shows that the gradient $\bm{g}$ is controlled by the following two components: (1) the direction to minimize $\mathcal{L}_t$ and its gradient norm $\|\nabla_{\bm{\theta}} \mathcal{L}_{t}(\bm{\theta})\|_2$. (2) the direction to minimize $\mathcal{L}_s$ and the corresponding norm $ \nabla_{\bm{\theta}} \mathcal{L}_{s}(\bm{\theta})$. Note that as illustrated in the \textit{stability-plasticity dilemma} \cite{lopez2017gradient}, the two loss terms con have potential conflicts both on the loss functions and the gradient norms. For example, when the angle between $\mathcal{L}_t$ and $\mathcal{L}_s$ is larger than $90^{\circ}$, the averaged gradient $\bm{g}^{ER-SAM}$ can lead to a suboptimal solution that either $\mathcal{L}_t$ or $\mathcal{L}_s$ is minimized. To address this issue, the proposed MGSER-SAM algorithm provides two regularization terms. First, instead of simply calculating the empirical loss, we leverage the model output logits to formulate an alternative loss function $\mathcal{\hat{L}}_{s}$ as:
\begin{equation}\label{Eq:mgser-logits}
	\mathcal{\hat{L}}_{s} = \mathbb{E}_{(\mathbf{x}, y) \sim \mathcal{B}} [l(f_\theta(\mathbf{x}), y)] + \mathbb{E}_{(\mathbf{x'}, \mathbf{z'}) \sim \mathcal{B}} [\|h_{\bm{\theta} (\mathbf{x'})} - \mathbf{z'} \|_2],
\end{equation}
where $\mathbf{z'}$ is the soft logits corresponding to the memory data $\mathbf{x'}$, where $h_{\bm{\theta}}(\mathbf{x'})$ ) represents the output of the final layer preceding the softmax layer, satisfying $\text{softmax}[h_{\bm{\theta}}(\mathbf{x'})] = f_{\bm{\theta}}(\mathbf{x'})$. We store the optimal $\mathbf{z'}$ as an attribute when learning the corresponding $\mathbf{x'}$ as part a new task. Consequently, the $l_2$ norm difference $\|h_{\bm{\theta} (\mathbf{x'})} - \mathbf{z'} \|_2$ quantifies the discrepancy between the logits post-learning new tasks and the optimal logits. As proved in \cite{hinton2015distilling,buzzega2020dark}, minimizing the second term in Eq.~\eqref{Eq:mgser-logits} effectively aligns with the optimization of the KL divergence, which minimizes the divergence between the probability distributions of the CL model before and after learning task $t$, in relation to the knowledge in $\mathcal{D}_s$. 

\begin{algorithm}[t!]
	\caption{MGSER-SAM}
	\begin{algorithmic}
	 \STATE Initialized $f_{\bm{\theta}}$, Memory buffer $\mathcal{M}$, Learning rate $\eta$.
	 \FOR {tasks $t = 1, \dots, T$}
	 \STATE Sample batch $\mathcal{B}_t$ for task $t$ from  $\mathcal{D}_t$.
	 \STATE Sample batch $\mathcal{B}_1=\{\mathbf{x},y\}$ from memory buffer $\mathcal{M}$.
	 \STATE Sample batch $\mathcal{B}_2=\{\mathbf{x'},\mathbf{z'}\}$ from memory buffer $\mathcal{M}$.
	 \STATE Given $\mathcal{B}_t$, and $\mathcal{B}_1$, compute the empirical loss $\mathcal{L}_{t}$ and $\mathcal{L}_{s}$.
	 \STATE Compute logits regularization term in Eq.~\eqref{Eq:mgser-logits}.
	 \STATE Gradient $\bm{g}^{MGSER-SAM}$ calculation in Eq.~\eqref{Eq:mgser-update}.
	 \STATE $\bm{\theta}_{t+1}   = \bm{\theta}_{t} - \eta \bm{g}^{MGSER-SAM}$.
	 \STATE Update memory buffer $\mathcal{M}$ with reservoir sampling.
	 \ENDFOR
	\end{algorithmic}
	\label{ALG:mg-ser}
 \end{algorithm}

Then, we further investigate the interplay between the SAM optimizer and the memory rehearsal technique commonly employed in ER-SAM. The ER method is conventionally understood to augment certain critical features in the CL model via experience rehearsal to reinfore prior knowledge in $\mathcal{D}_s$. However, as pointed out in \cite{qu2023prevent}, exclusively concentrating on distinct features in the memory $\mathcal{M}$ may result in a skewed representation of knowledge in $\mathcal{D}_s$. Conversely, introducing perturbations across the entire ER process vis SAM could diminish the intrinsic influence of $\mathcal{M}$. To mitigate this, we utilize the empirical loss of $\mathcal{\hat{L}}_{s}$  a secondary regularization term. This serves to guide the update direction of the sharpness-aware minimization optimizer, ensuring a more balanced and comprehensive learning process. The update mechanism for the MGSER-SAM model is articulated as follows:
\begin{equation}\label{Eq:mgser-update}
	\begin{split}
		& \bm{g}^{MGSER-SAM}  = \nabla_{\bm{\theta}} \mathcal{\hat{L}}_{total} (\bm{\theta}) |_{\bm{\theta}+ \bdelta^{\star}} + \nabla_{\bm{\theta}} \mathcal{\hat{L}}_{s} (\bm{\theta}) |_{\bm{\theta}} , \\
		& \mathcal{\hat{L}}_{total}    = \mathcal{L}_{t} + \mathcal{\hat{L}}_{s},  \quad  \bm{\theta}_{t+1}   = \bm{\theta}_{t} - \eta \bm{g}^{MGSER-SAM}.  \\
	\end{split}
\end{equation}

\subsection{Discussion}
To more effectively illustrate the learning process of our proposed MGSER-SAM algorithm, we have detailed the pseudo-code in Algorithm~\ref{ALG:mg-ser}. This paper introduces several key contributions to enhance the performance of experience replay, particularly through the  generalization capabilities enabled by sharpness-aware minimization. Our first contribution is the introduction of the ER-SAM algorithm, which applies SAM directly to experience replay (ER). It is important to note that the established SAM optimizer can be viewed as a flatness component, which is able to seamlessly integrated into various ER-based methods, such as Dark Experience Replay (DER++ \cite{buzzega2020dark}). We will delve deeper into this integration in the experimental section of this paper.

\begin{table}[b]
	\caption{Numerical details of introduced benchmarks in this paper for three scenarios in CL. $N_{task}$ is the total number of tasks, and $N_{c}$ and $N_{train}$ denote the number of training data points and the classification class number per each task.}
	\begin{adjustbox}{width=0.95\columnwidth, center}
		\begin{tabular}{*{5}{c}}
			\toprule
			Dataset  & Scenario& $N_{train}$& $N_{task}$ & $N_{c}$  \\  
			\midrule
			S-MNIST \cite{lecun1998gradient}& Task-IL and Class-IL& $12,000$ & $5$ & $ 2$ \\
			S-CIFAR10 \cite{krizhevsky2009learning}& Task-IL and Class-IL& $10,000$ & $5$  & $ 2$ \\
			S-CIFAR100 \cite{krizhevsky2009learning} &Task-IL and Class-IL &$5,000$ & $10$  & $ 10$ \\
			S-TinyImageNet \cite{Le2015TinyIV} &Task-IL and Class-IL   & $5,000$ & $10$ & $20$ \\
			P-MNIST \cite{kirkpatrick2017overcoming}& Domain-IL& $1000$ & $20$  & $ 10$ \\
			R-MNIST \cite{lopez2017gradient} &Domain-IL &$1000$ & $20$  & $10$ \\
			\bottomrule
		\end{tabular}
	\end{adjustbox}

	\label{Table:data}
\end{table}

Secondly, we explore a conjecture regarding the potential conflict between weight perturbation directions arising from the current task and the memory in terms of both gradient norm and empirical loss. To address this conflict, we propose the MGSER-SAM algorithm, which utilizes soft logits and memory gradient directions as regularization terms. These terms are designed to effectively guide the update direction in sharpness-aware minimization. It should be emphasized that the two loss terms presented in Eq.~\eqref{Eq:mgser-logits} offer the potential for further investigation, particularly regarding the introduction of a weighted hyperparameter. This parameter could effectively balance the significance of the two terms. However, for the purposes of this study, we have chosen to set the same weight for  simplicity. It is important to clarify that the fine-tuning of the balance in $\mathcal{\hat{L}_s}$ is not the focus of this research.

Compared to ER, the computational overhead of MGSER-SAM and ER-SAM is primarily due to the additional back-propagation operation required for calculating $\bdelta^{\star}$. While the extra batch sampling for $\mathcal{B}_2$ for soft logits, as indicated at Line~5, is relatively insignificant. Consequently, the computational cost of ER-SAM and MGSER-SAM is approximately twice that of standard ER. Nevertheless, given that both ER-SAM and MGSER-SAM are memory-based CL approaches, their computational costs are relatively minor compared to memory access costs. Furthermore, the MGSER-SAM algorithm does not incur any additional computational expense for calculating model differences, rendering it computationally efficient compared to other studies such as MIR \cite{aljundi2019online}.

\section{Experiments}
\label{sec:experiment}
\subsection{Experimental Setup}


To evaluate the proposed MGSER-SAM algorithm and the flatness SAM optimizer, we compared them with the representative algorithms under all the three CL scenarios on multiple benchmarks \cite{van2019three}. The experimental environment in this paper is developed based on the popular opensource continual learning library\footnote{\href{https://github.com/aimagelab/mammoth}{\textcolor{blue}{https://github.com/aimagelab/mammoth\cite{boschini2022class}}}}.

\subsubsection{Benchmarks}
The benchmarks introduced in this paper are tailored for the three CL scenarios, including task-IL, class-IL, and domain-IL, where the numerical details are introduced in Table.~\ref{Table:data}. For the task-IL and class-IL benchmarks, we have selected S-MNIST, S-CIFAR10, S-CIFAR100, and S-TinyImageNet. For the domain-IL benchmarks, we have selected P-MNIST and R-MNIST.

\subsubsection{Baselines}
We compare the proposed MGSER-SAM algorithm with the following representative baselines in recent literature: \textbf{LWF} \cite{li2017learning}, PNN \cite{rusu2016progressive}, \textbf{SI} \cite{zenke2017continual},  \textbf{oEWC} \cite{schwarz2018progress}, \textbf{ER} \cite{rolnick2018experience}, and \textbf{DER++} \cite{buzzega2020dark}. We also evaluate the performance of two settings: \textbf{Online:} the learner is trained in CL without any forgetting mitigation; \textbf{Joint:} the tasks are trained jointly as one dataset. Moreover, to evaluate the adaptation of the SAM optimizer, we also compare MGSER-SAM with \textbf{ER-SAM} and \textbf{DER++-SAM} that are the SAM variants of ER and DER++.

\begin{table*}[t!]
    \caption{Three scenarios CL results for benchmarks S-MNIST, S-CIFAR10, S-CIFAR100, S-TinyImageNet, P-MNIST, and R-MNIST. We report the Acc metrics (higher is better). The results are splitted into different categories via the horizontal lines: the non memory-replay baselines, and the different size of memory buffer. (Method with * indicates the results for  S-CIFAR10, S-TinyImageNet, P-MNIST, and R-MNIST are reported in \cite{buzzega2020dark}, and the best performance is denoted with \textbf{bold} fonts).}
	\begin{adjustbox}{width=0.99\textwidth, center}
		\begin{tabular}{*{12}{c}}
			\toprule
			 &\multicolumn{1}{c}{}&  \multicolumn{2}{c}{S-MNIST}   & \multicolumn{2}{c}{S-CIFAR10} & \multicolumn{2}{c}{S-CIFAR100} &  \multicolumn{2}{c}{S-TinyImageNet} & {P-MNIST} & {R-MNIST}  \\
			{Buffer Size}&Method & Class-IL & Task-IL & Class-IL & Task-IL & Class-IL & Task-IL & Class-IL & Task-IL & Domain-IL& Domain-IL  \\
			\midrule
			\multirow{ 2}{*}{-}&Online$^*$ & $19.25$ &$88.49$ & $19.62$ & $61.23$ &$69.96 $ & $35.08$ & $7.92$ &$18.31$ &$40.70$ & $67.66$ \\
			&Joint$^*$  & $92.53$ &$98.72$ & $92.20$ &$98.31$ &$9.09$ & $90.94$ & $59.99$ &$82.04$& $94.33 $ & $95.76$ \\
			\midrule
			\multirow{ 4}{*}{-}&LWF$^*$ & $19.24$ &$ 97.55$ & $19.61$ & $63.29$ &$8.9$ & $23.29$ & $8.46$ &$15.85$& N/A &N/A \\
			&PNN$^*$ & N/A &$97.99$ & N/A & $95.13$ &N/A & $57.13$ & N/A  &$67.84$& N/A &N/A \\	
			&SI$^*$ & $19.29 $ &$91.40$ & $19.48$ & $68.05$ &$9.08$ & $26.75$ & $6.58$ &$36.32$& $65.86$ &$71.91$  \\
			&oEWC$^*$ & $19.29$ &$97.49$ & $19.49$ & $68.29$ &$5.97$ & $17.77$ & $7.58$ &$19.20$& $75.79 $ &$ 77.35$  \\			

            \midrule
            \multirow{ 5}{*}{400}&ER & $86.67 $ &$ 98.13$ & $55.51$ & $92.58$ &$15.41$ & $61.45$ & $8.56$ &$39.06$& $73.02$ &$83.00$  \\	
			&ER-SAM & $87.53 $ & \bm{$98.58$} & $47.73$ & $86.59$ &$15.89$ & $61.16$ & $8.44$ &$33.69$& $81.28$  & $87.76$ \\
            &DER++ & $78.66 $ &$ 98.16$ & $68.38$ & $93.26$ &$23.91$ & $65.95$ & $11.23$ &$39.41$& $83.56$ &$ 90.08$  \\	
			&DER++-SAM & $87.31$ & $98.05$ & $61.37$ & $89.52$ &$28.83$ & $69.57$ & $14.52$ &$45.31$& $84.18$  & \bm{$93.73$} \\
            &MGSER-SAM & \bm{$88.31$} & $98.34$ & \bm{$72.03$} & \bm{$93.41$} &\bm{$29.98$} & \bm{$70.08$} & \bm{$14.53$} &\bm{$45.71$}& \bm{$85.19$}  & $92.87$\\
            \midrule
            \multirow{ 5}{*}{1000}&ER & $90.06 $ &$ 98.01$ & $63.11$& $93.16$ &$22.38$ & $69.80$ & $9.81$ &$46.05$& $78.55$ &$ 87.72$  \\	
			&ER-SAM & $89.26 $ & $98.48$ & $62.21$ & $88.89$ &$22.84$ & $70.23$ & $10.51$ &$49.65$& $85.02$  & $91.28$  \\
            &DER++ & $79.42$ &$ 98.13$ & $74.69$ & $94.02$ &$40.15$ & $76.90$ & $18.03$ &$51.86$& $86.21$ &$ 90.41$  \\	
			&DER++-SAM & $90.25$ & $98.40$ & $75.76$ & $94.86$ &\bm{$41.87$} & $76.18$ & $20.15$ &$53.54$& $89.38$  & \bm{$93.79$}  \\
            &MGSER-SAM & \bm{$91.75 $} & \bm{$98.52$} & \bm{$78.51$} & \bm{$94.88$} &$40.20$& \bm{$77.03$} & \bm{$21.22$} & \bm{$54.01$}& \bm{$89.91$}  & $92.91$\\

            \midrule

            \multirow{ 5}{*}{2000}&ER & $89.49 $ &$98.04$ & $75.46$ & $95.23$& $31.11$ &$75.09$  & $14.14$ &$53.45$& $80.27$ &$89.43$  \\	
			&ER-SAM & $89.70$ & $98.58$ & $71.63$ & $93.37$  & $33.92$ & $75.23$  &$13.95$ &$47.25$& $86.34$  & $92.62$  \\
            &DER++ & $79.33 $ &$98.15$ & $77.91$ & $94.38$ &$47.15$ & $81.11$ & $28.61$ &$61.30$& $87.56$ &$ 91.29$  \\	
			&DER++-SAM & $90.13$ & $98.59$ & $80.99$& $95.74$ &\bm{$51.0$} & \bm{$81.89$} & $29.32$&$61.55$& $91.00$  & \bm{$93.94$}  \\
            &MGSER-SAM & \bm{$93.29$} & \bm{$98.88$} & \bm{$81.99$} & \bm{$95.83$} &$49.92$ & $79.91$  & \bm{$30.38$} &\bm{$63.05$}& \bm{$91.49$}  & $93.76$\\
			\bottomrule
		\end{tabular}
	\end{adjustbox}

	\label{Table:analysis}
\end{table*}

\subsubsection{Evaluation metrics}
To provide a fair comparison with the compared CL baselines, we train all the models in the paper with the same hyper-parameters and computational resources for each benchmark. For S-MNIST, P-MNIST, and R-MNIST, we train a two-lay MLP with 400 hidden nodes \cite{lopez2017gradient}. While for S-CIFAR10, S-CIFAR100, and S-TinyImageNet, we use a Resnet-18 \cite{he2016deep}. For the memory replay based baselines, the batch size $\mathcal{B}_t$,  $\mathcal{B}_1 $and $\mathcal{B}_2$ are set to the same. The results in the paper are reported with the average of fives repetitions with different random seeds. We measure the performance with the following metrics as defined in \cite{lopez2017gradient,chaudhry2018efficient}. Note that for CL with $T$ tasks, the testing accuracy is evaluated after each task, which can be introduced as a result matrix $R \in \mathbb{R}^{T \times T}$ where $R_{i,j}$ is the accuracy of $j$-th task after learning task $i$. Let $F_i$ be the best testing accuracy for the $i$-th task, we focus the follows: (1) average accuracy (ACC): $\frac{1}{T}\sum_{i=1}^{T}R_{T,i} $; (2) forgetting (Forget): $\frac{1}{T-1}\sum_{i=1}^{T-1} R_{T,i} - F_i $.

\subsection{Results}
\subsubsection{Performance analysis}
Table.~\ref{Table:analysis} shows the performance of compared CL baselines against the six introduced benchmarks. The results show that the proposed MGSER-SAM algorithm achieves the best overall performance.
\begin{figure*}[ht]
	\centering
	\begin{subfigure}{1\columnwidth}
		\includegraphics[width=\columnwidth]{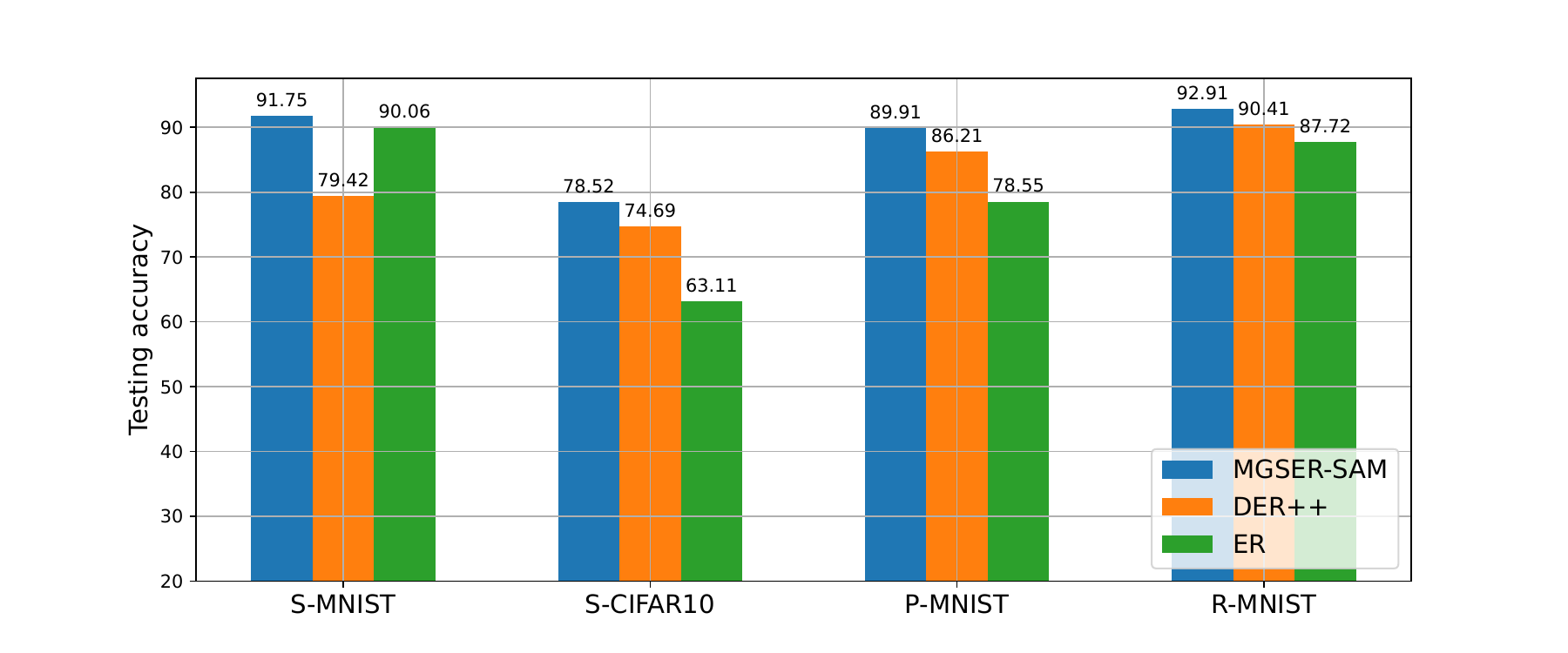}
		\caption{ACC}
		\label{fig:bars_acc}
	\end{subfigure}
	\begin{subfigure}{1\columnwidth}
		\includegraphics[width=\columnwidth]{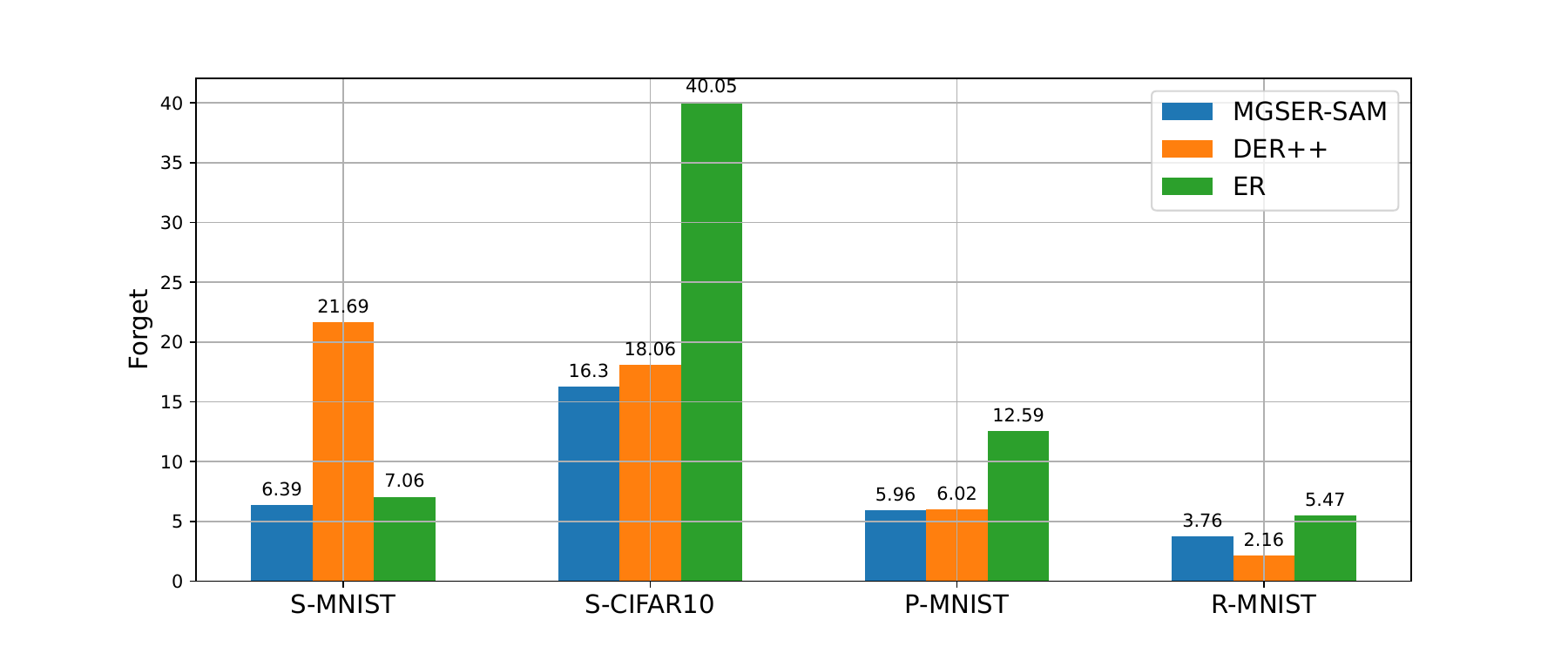}
		\caption{Forget}
		\label{fig:bars_forget}
	\end{subfigure}	
	\caption{Results for the compared baselines on ACC and Forget metrics over S-MNIST, S-CIFAR10, P-MNIST, and R-MNIST benchmarks: \textbf{ACC}: higher is better; \textbf{Forget}: lower is better.}
    \label{fig:bars}
\end{figure*}

Specifically, for S-MNIST with $M = 2000$, MGSER-SAM achieves $93.29\%$ testing accuracy in class-IL, which is $4.2\%$ and $17.6\%$ higher than the ER and DER++ methods respectively. For the S-CIFAR10 benchmark with $M=1000$, MGSER-SAM obtains $78.51\%$ testing accuracy in class-IL, which is $24.4\%$ higher than the ER method. For the S-CIFAR100 benchmark with $M=400$, MGSER-SAM achieves $70.08\%$ testing accuracy in task-IL, which is $6.3\%$ higher than the DER++ method. For the S-TinyImageNet benchmark with $M=1000$, MGSER-SAM obtains the best $21.22\%$ and $54.01\%$ testing accuracy in class-IL and task-IL, respectively. Additionally, for the domain-IL scenario, MGSER-SAM achieves the best performance on P-MNIST. 

We notice several interesting phenomenons: (i) Though ER-SAM obtains overall better performance than ER, it performs poorly on the S-CIFAR10 benchmark, especially when the buffer size is small; (ii) The DER++-SAM shows a significant performance improvement over DER++ on the task-IL and domain-IL scenarios. Meanwhile, we monitor the Forget metrics of MGSER-SAM on multiple benchmarks, compared with ER and DER++ in Figure.~\ref{fig:bars}. The results show that MGSER-SAM can most effectively address the catastrophic forgetting problem. Though it has a higher Forget on R-MNIST, it still achieves the best performance on the ACC metrics.

\subsubsection{Investigation during CL process}
\begin{figure*}[ht]
    \centering
    \begin{subfigure}{0.66\columnwidth}
    \includegraphics[width = 1\columnwidth]{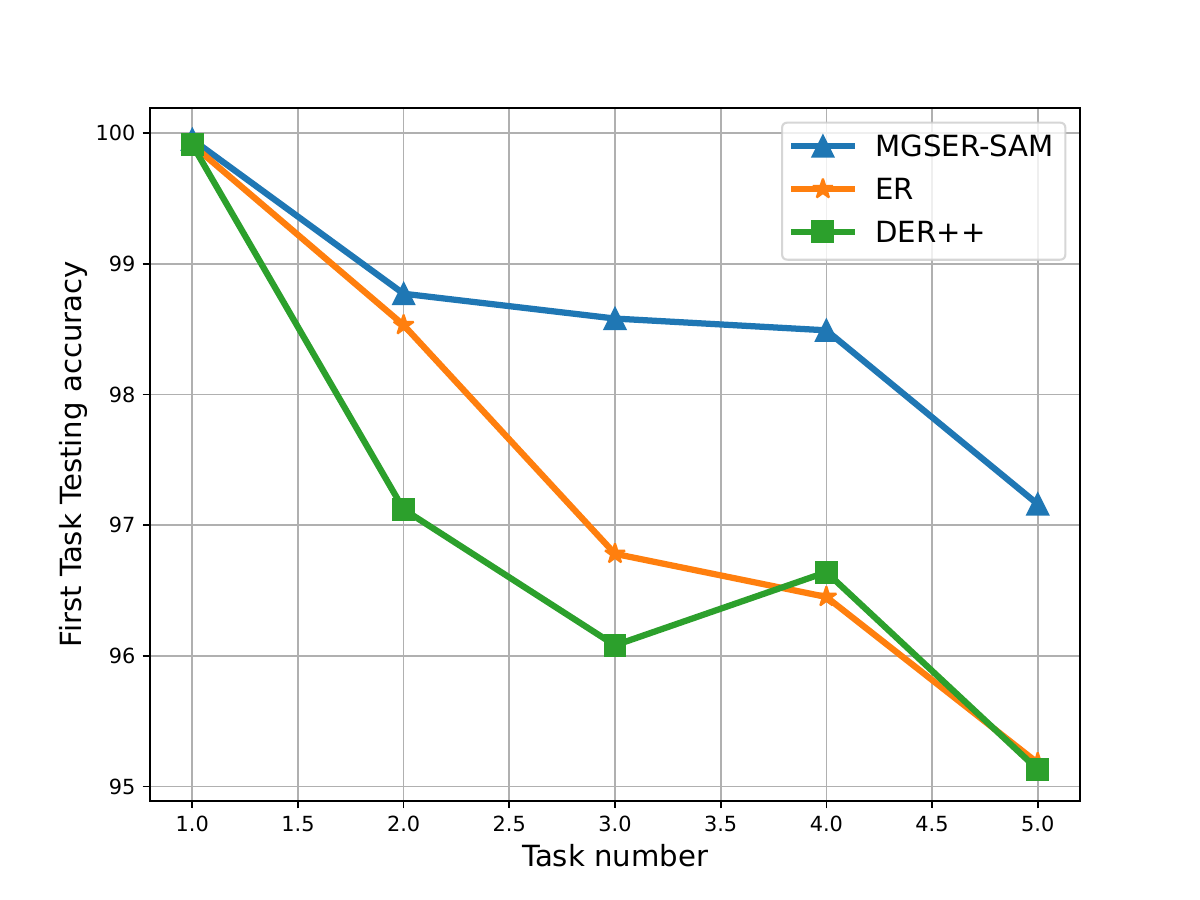}
    \caption{S-MNIST}
    \label{fig:mnist_first_acc}
    \end{subfigure}
    \begin{subfigure}{0.66\columnwidth}
    \includegraphics[width = 1\columnwidth]{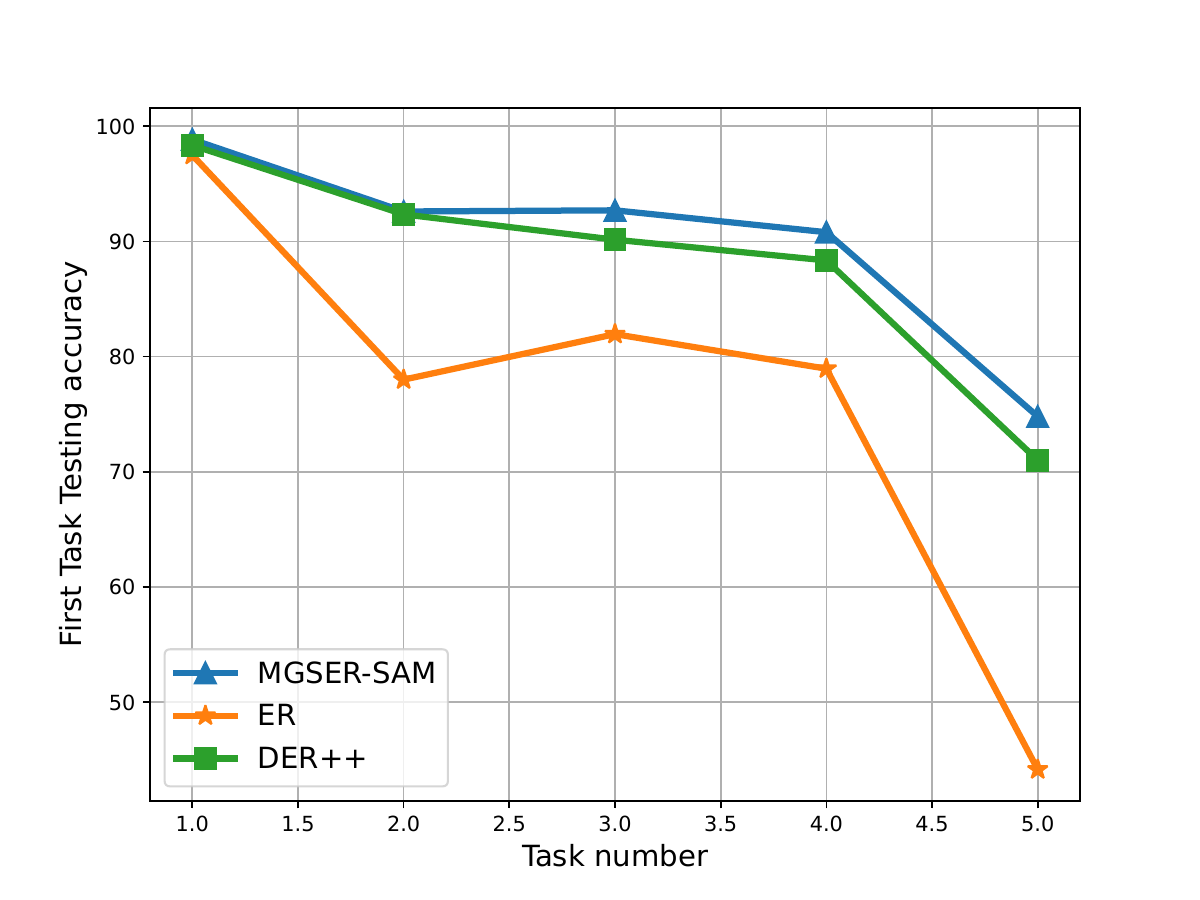}
    \caption{S-CIFAR10}
    \label{fig:cifar10_first_acc}
    \end{subfigure}
    \begin{subfigure}{0.66\columnwidth}
    \includegraphics[width = 1\columnwidth]{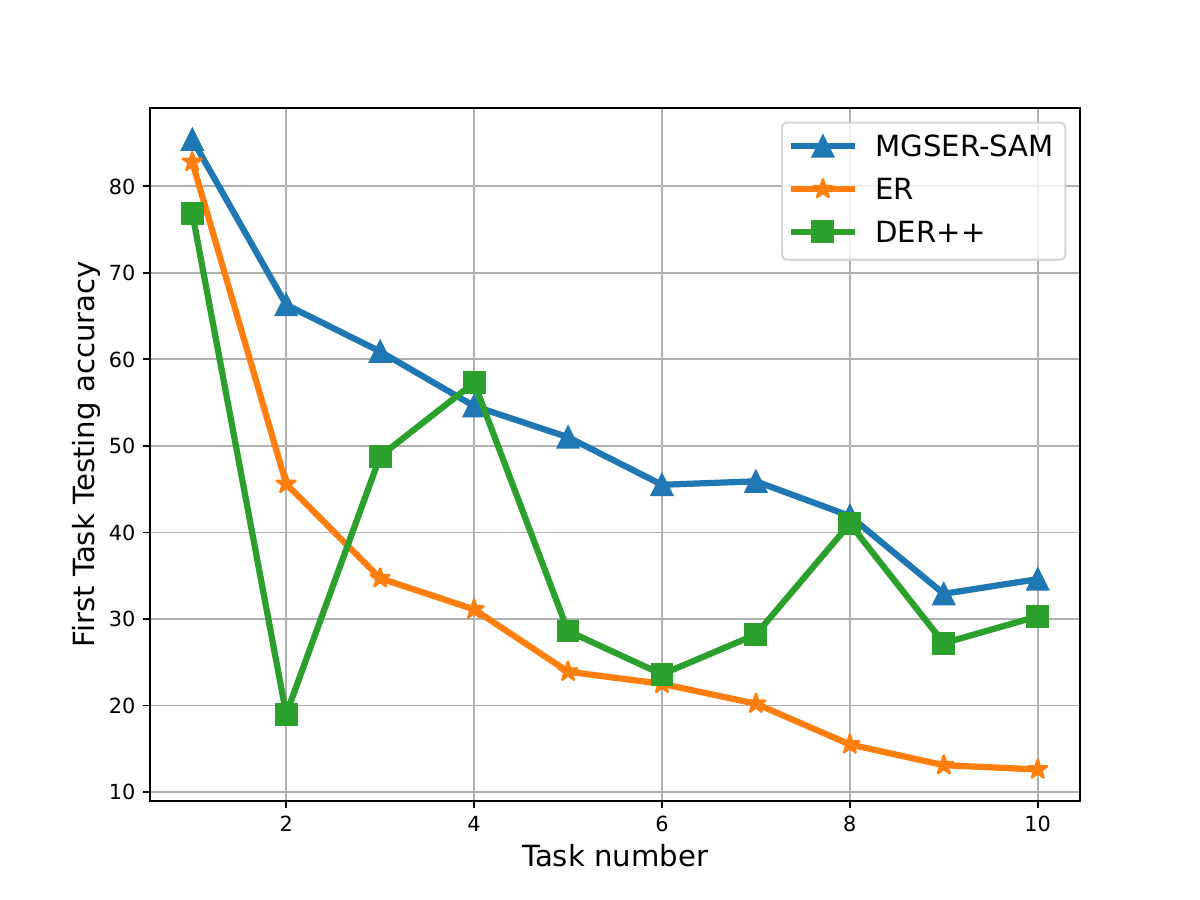}
    \caption{S-CIFAR100}
    \label{fig:cifar100_first_acc}
    \end{subfigure}
    \caption{Results for the compared baselines on the change of first task testing accuracy during the CL learning process.}
    \label{fig:first_acc}
\end{figure*}
\begin{figure*}[ht]
	\centering
	\begin{subfigure}{1\columnwidth}
		\includegraphics[width=\columnwidth]{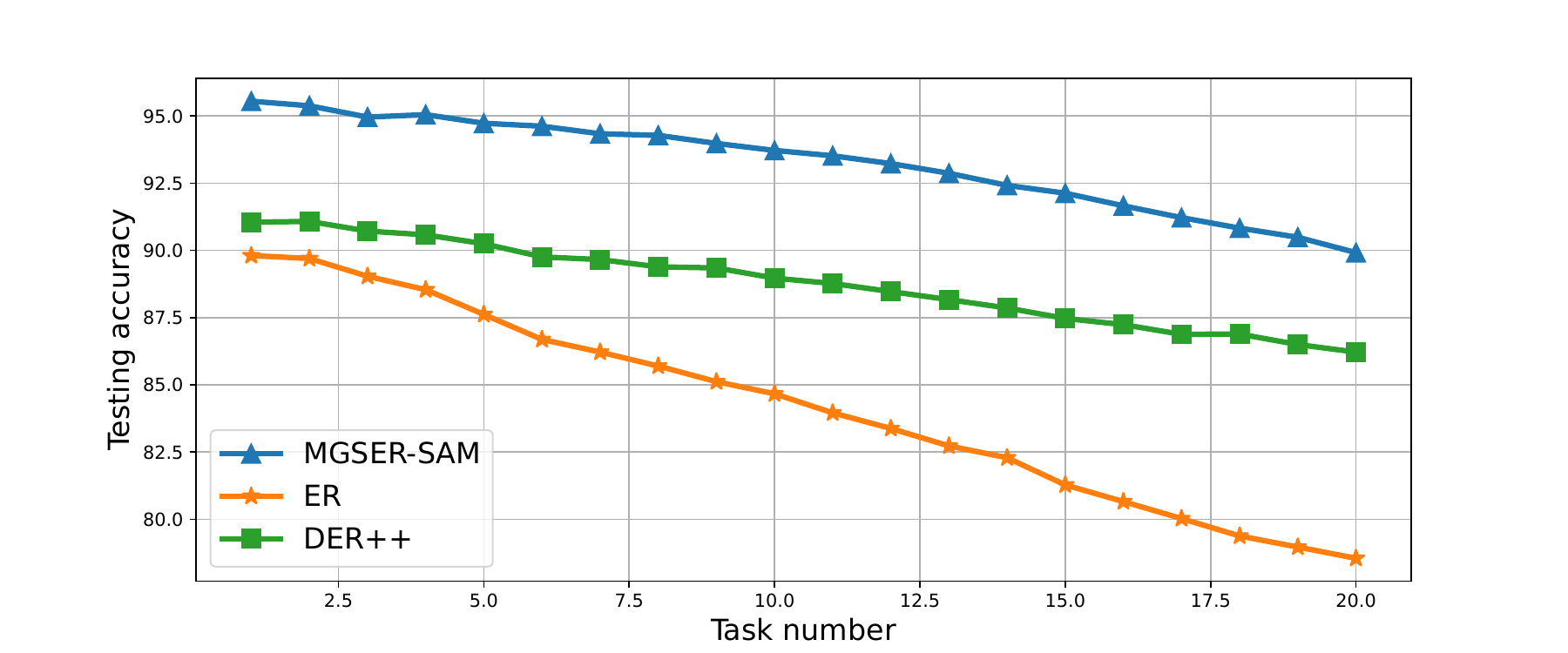}
		\caption{P-MNIST  }
		\label{fig:task_pmnist}
	\end{subfigure}
	\begin{subfigure}{1\columnwidth}
		\includegraphics[width=\columnwidth]{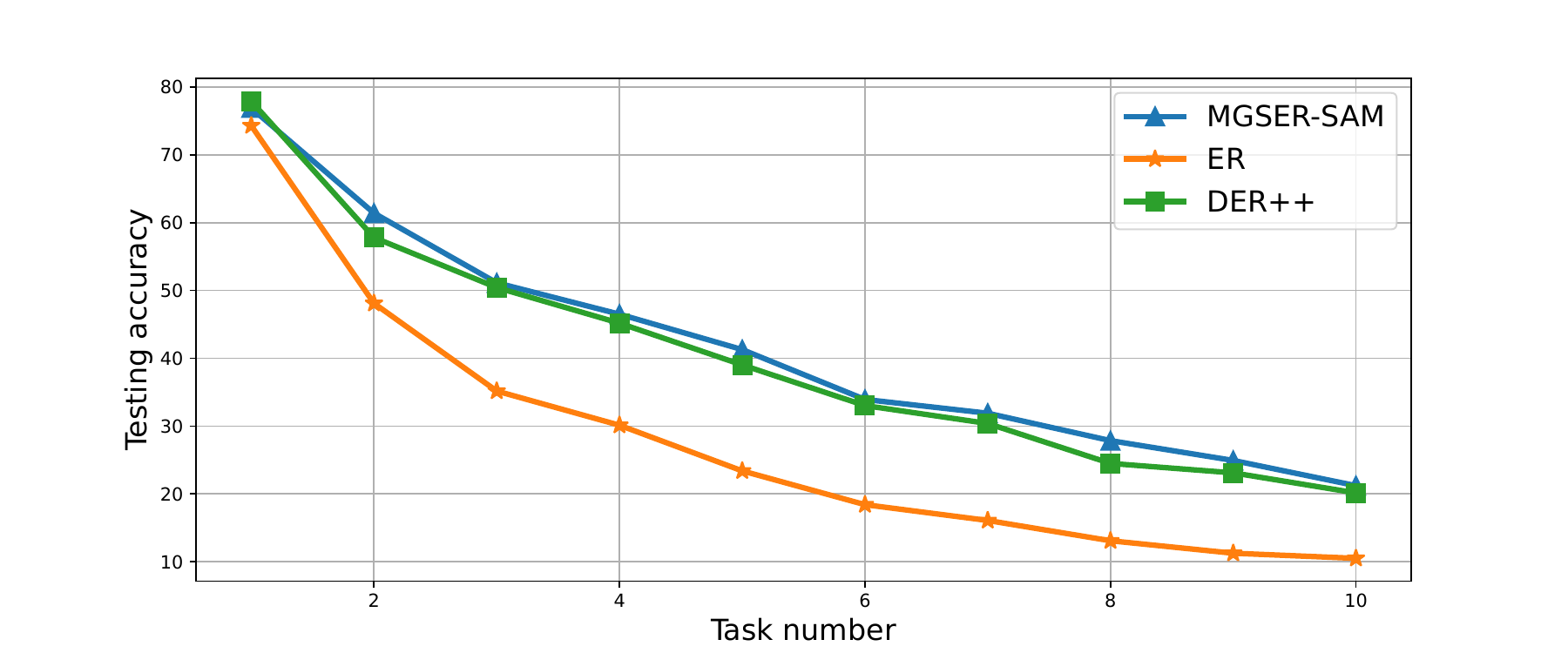}
		\caption{S-TinyImageNet}
		\label{fig:task_tiny}
	\end{subfigure}	
	\caption{The ACC of incrementally learning all tasks during class-IL CL.}
	\label{fig:task_long}
\end{figure*}
\begin{figure*}[ht]
    \centering
    \begin{subfigure}{0.66\columnwidth}
    \includegraphics[width = 1\columnwidth]{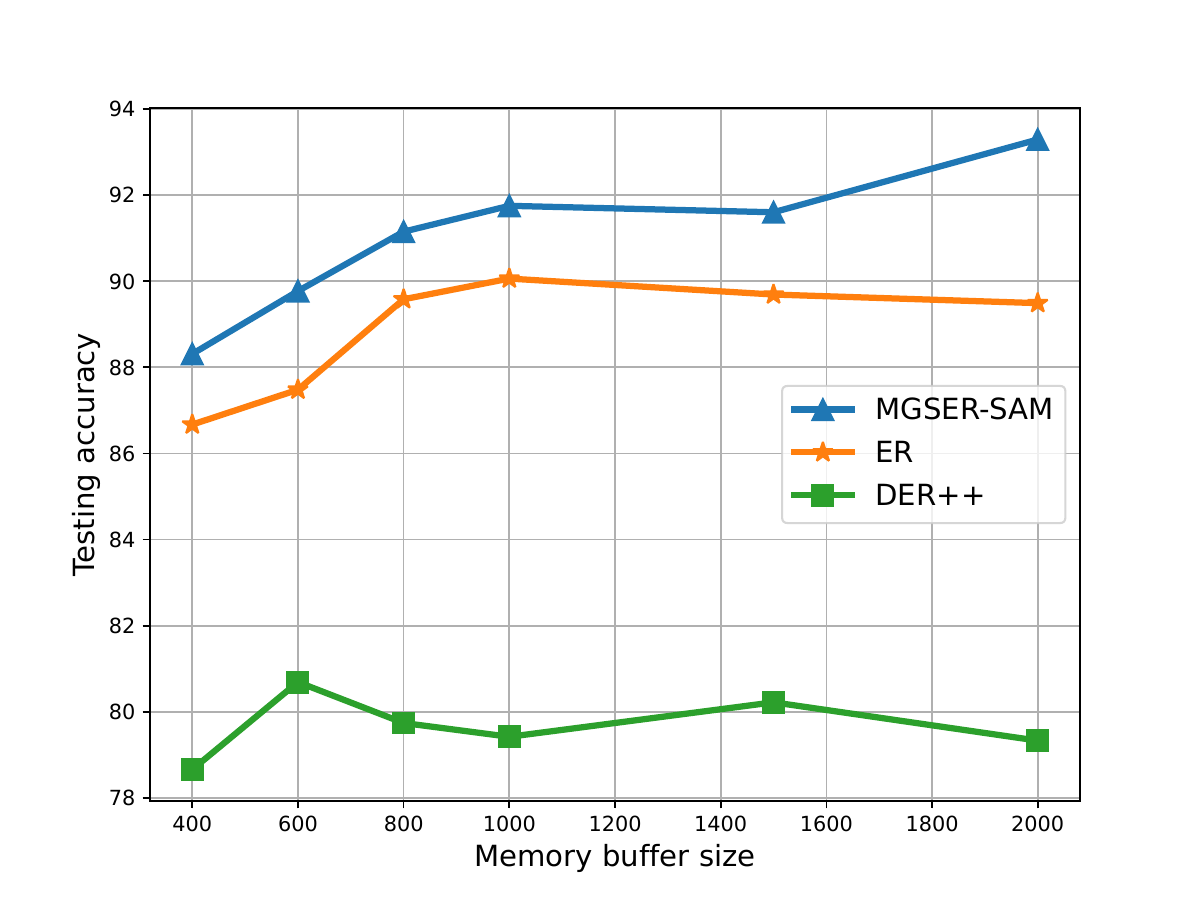}
    \caption{S-MNIST}
    \label{fig:mnist_mem_acc}
    \end{subfigure}
    \begin{subfigure}{0.66\columnwidth}
    \includegraphics[width = 1\columnwidth]{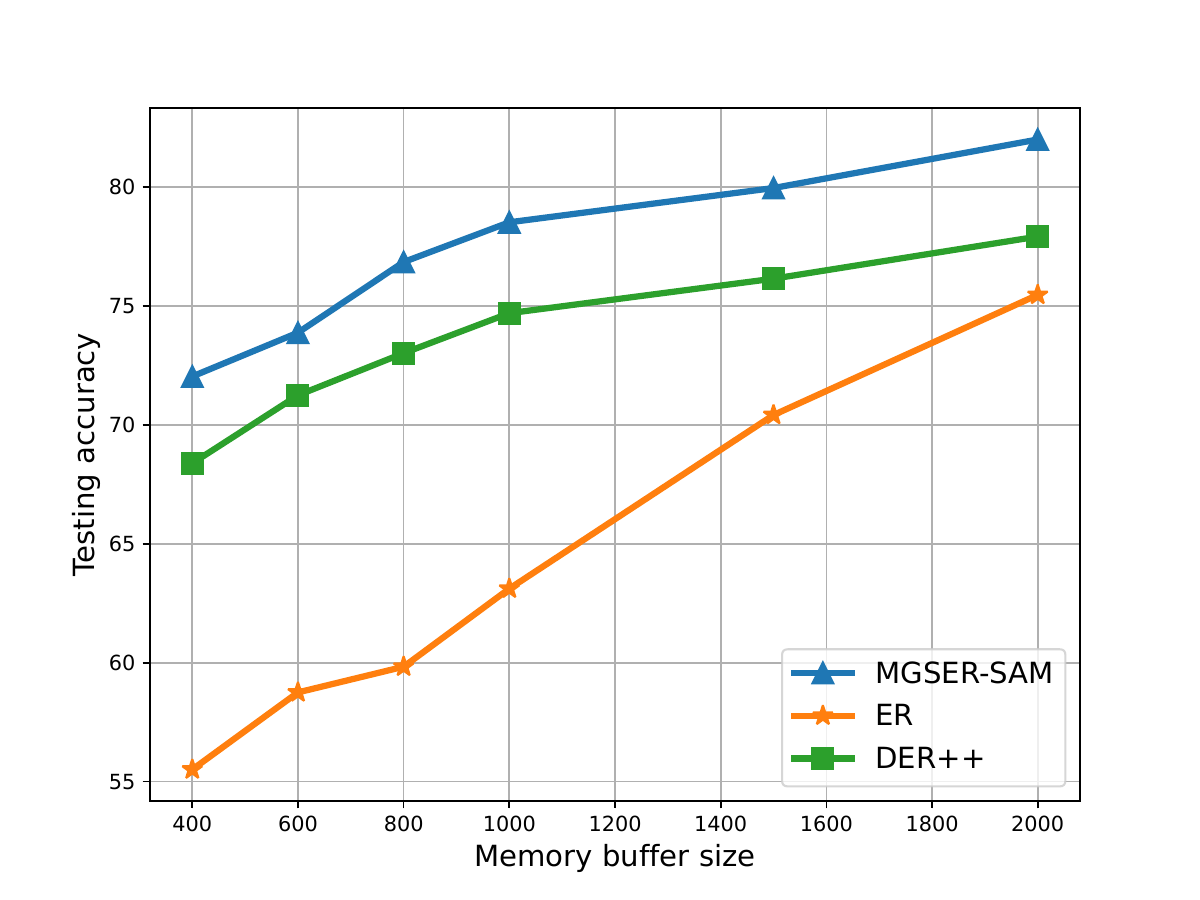}
    \caption{S-CIFAR10}
    \label{fig:fmnist_mem_acc}
    \end{subfigure}
    \begin{subfigure}{0.66\columnwidth}
    \includegraphics[width = 1\columnwidth]{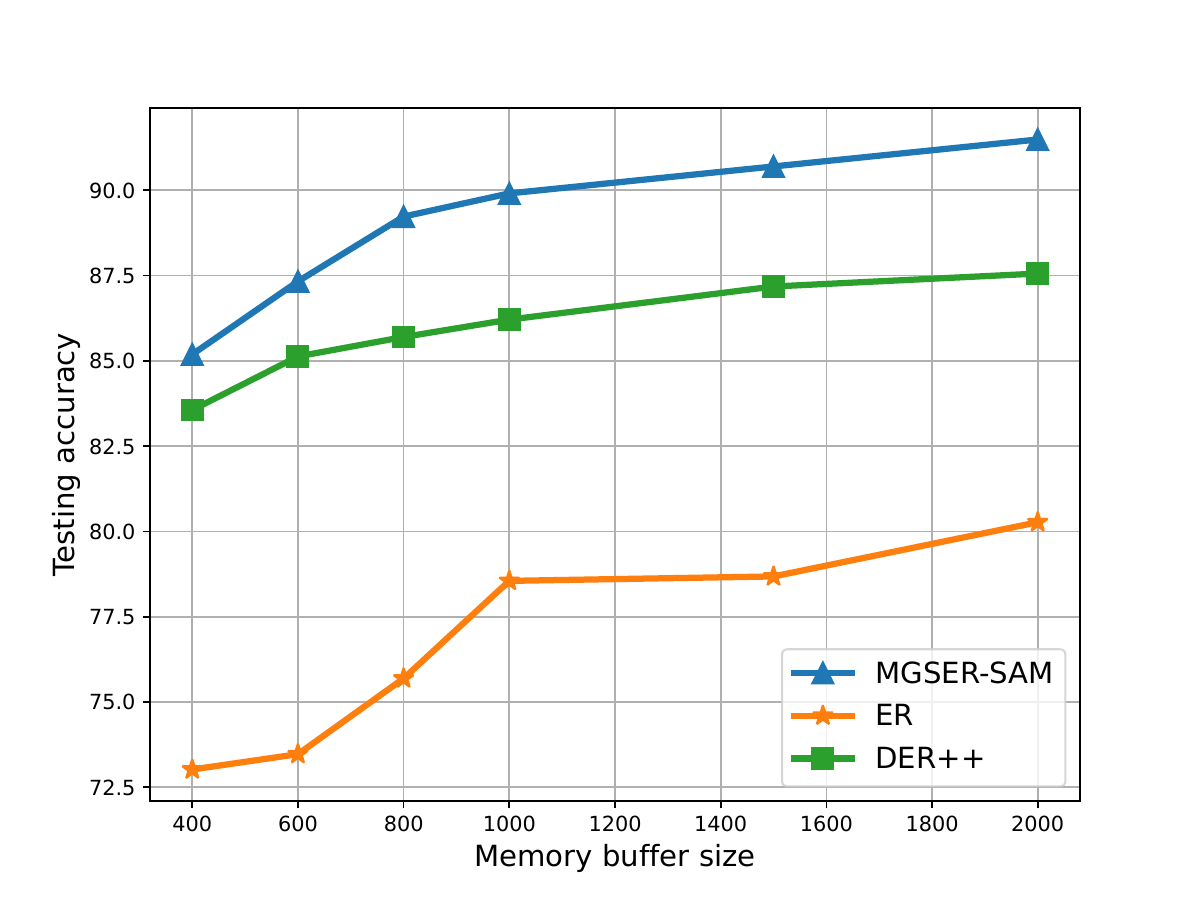}
    \caption{P-MNIST}
    \label{fig:cifar_mem_acc}
    \end{subfigure}
    \caption{The impact of different memory size in class-IL CL over ACC.}
    \label{fig:mem}
\end{figure*}

We then study the performance of compared baselines during the CL process. In Figure.~\ref{fig:first_acc}, we monitor the testing accuracy of the first task in S-MNIST, S-CIFAR10, and S-CIFAR100 to present how the forgetting evolves as new tasks are sequentially learned. The results show that MGSER-SAM achieves the overall highest testing accuracy on the first task after learning each tasks in all considered benchmarks. For example, the first task accuracy of MGSER-SAM decreases $24.05\%$ on S-CIFAR10, which is $54.92\%$ and $12.06\%$ lower than the loss on ER and DER++. 

The results in Figure.~\ref{fig:task_long} show the averaged ACC of compared baselines on P-MNIST and S-TinyImageNet. The results show that MGSER-SAM achieves the highest ACC on both benchmarks after the learning of each task. Specifically, for P-MNIST, the ACC of MGSER-SAM after learning 20 tasks is $89.92\%$, which is higher than the ACC of ER after learning the first task, indicating its superiority.

\subsubsection{Impacts of memory capacity} 

We then study ACC performance with different memory buffer size where $M \in [400, 2000]$. The results shown in Figure.~\ref{fig:mem} show that MGSER-SAM achieves the best ACC performance against class-IL CL on S-MNIST, S-CIFAR10, and P-MNIST benchmarks with each considered memory buffer size. We also observed that as $M$ increases, the performance of all compared CL baselines becomes better. Comparing to the overall second-best performance of DER++, the advantage of MGSER-SAM becomes more significant as $M$ increases. For example, on P-MNIST, the ACC difference Between MGSER-SAM and DER++ increases from $1.6\%$ to $3.93\%$.

\section{Conclusion}
\label{sec:conclusion}
The challenge of catastrophic forgetting stands as a major obstacle in the advancement of continual learning. This paper introduces a pioneering memory replay-based algorithm, MGSER-SAM, specifically designed to mitigate this challenge by enhancing the generalization capabilities of continual learning (CL) models. We commence by presenting the SAM optimizer, a component focused on optimizing flatness, which can be seamlessly integrated into existing Experience Replay-based frameworks such as ER and DER++. Building upon this, the MGSER-SAM algorithm is proposed, adeptly addressing the potential conflicts between weight perturbation directions in the current task and previously learned memories within the SAM optimizer. By introducing the soft logits and memory gradient direction alignment, MGSER-SAM can effectively minimize multiple training loss terms concurrently during the CL process. Our experimental evaluations demonstrate the effectiveness of SAM optimizer and the superiority of the proposed MGSER-SAM against compared baselines across various benchmarks in all three CL scenarios.

\section{Acknowledgement}

This research was funded by US National Science Foundation (NSF), Award IIS 2325863.





\bibliographystyle{IEEEbib}
\bibliography{refs}

\begin{thebibliography}{10}

\bibitem{krizhevsky2012imagenet}
Alex Krizhevsky, Ilya Sutskever, and Geoffrey~E Hinton,
\newblock ``Imagenet classification with deep convolutional neural networks,''
\newblock {\em Advances in neural information processing systems}, vol. 25, pp. 1097--1105, 2012.

\bibitem{li2021lomar}
Xingyu Li, Zhe Qu, Shangqing Zhao, Bo~Tang, Zhuo Lu, and Yao Liu,
\newblock ``Lomar: A local defense against poisoning attack on federated learning,''
\newblock {\em IEEE Transactions on Dependable and Secure Computing}, 2021.

\bibitem{li2023fedlga}
Xingyu Li, Zhe Qu, Bo~Tang, and Zhuo Lu,
\newblock ``Fedlga: Toward system-heterogeneity of federated learning via local gradient approximation,''
\newblock {\em IEEE Transactions on Cybernetics}, 2023.

\bibitem{goodfellow2013empirical}
Ian~J Goodfellow, Mehdi Mirza, Da~Xiao, Aaron Courville, and Yoshua Bengio,
\newblock ``An empirical investigation of catastrophic forgetting in gradient-based neural networks,''
\newblock {\em arXiv preprint arXiv:1312.6211}, 2013.

\bibitem{kirkpatrick2017overcoming}
James Kirkpatrick, Razvan Pascanu, Neil Rabinowitz, Joel Veness, Guillaume Desjardins, Andrei~A Rusu, Kieran Milan, John Quan, Tiago Ramalho, Agnieszka Grabska-Barwinska, et~al.,
\newblock ``Overcoming catastrophic forgetting in neural networks,''
\newblock {\em Proceedings of the national academy of sciences}, vol. 114, no. 13, pp. 3521--3526, 2017.

\bibitem{zenke2017continual}
Friedemann Zenke, Ben Poole, and Surya Ganguli,
\newblock ``Continual learning through synaptic intelligence,''
\newblock in {\em International Conference on Machine Learning}. PMLR, 2017, pp. 3987--3995.

\bibitem{rusu2016progressive}
Andrei~A Rusu, Neil~C Rabinowitz, Guillaume Desjardins, Hubert Soyer, James Kirkpatrick, Koray Kavukcuoglu, Razvan Pascanu, and Raia Hadsell,
\newblock ``Progressive neural networks,''
\newblock {\em arXiv preprint arXiv:1606.04671}, 2016.

\bibitem{lopez2017gradient}
David Lopez-Paz and Marc'Aurelio Ranzato,
\newblock ``Gradient episodic memory for continual learning,''
\newblock in {\em Proceedings of the 31st International Conference on Neural Information Processing Systems}, 2017, pp. 6470--6479.

\bibitem{rolnick2018experience}
David Rolnick, Arun Ahuja, Jonathan Schwarz, Timothy Lillicrap, and Gregory Wayne,
\newblock ``Experience replay for continual learning,''
\newblock {\em Advances in Neural Information Processing Systems}, vol. 32, 2019.

\bibitem{van2019three}
Gido~M Van~de Ven and Andreas~S Tolias,
\newblock ``Three scenarios for continual learning,''
\newblock {\em arXiv preprint arXiv:1904.07734}, 2019.

\bibitem{vitter1985random}
Jeffrey~S Vitter,
\newblock ``Random sampling with a reservoir,''
\newblock {\em ACM Transactions on Mathematical Software (TOMS)}, vol. 11, no. 1, pp. 37--57, 1985.

\bibitem{keskar2017on}
Nitish~Shirish Keskar, Dheevatsa Mudigere, Jorge Nocedal, Mikhail Smelyanskiy, and Ping Tak~Peter Tang,
\newblock ``On large-batch training for deep learning: Generalization gap and sharp minima,''
\newblock in {\em International Conference on Learning Representations}, 2017.

\bibitem{chaudhari2019entropy}
Pratik Chaudhari, Anna Choromanska, Stefano Soatto, Yann LeCun, Carlo Baldassi, Christian Borgs, Jennifer Chayes, Levent Sagun, and Riccardo Zecchina,
\newblock ``Entropy-sgd: Biasing gradient descent into wide valleys,''
\newblock {\em Journal of Statistical Mechanics: Theory and Experiment}, vol. 2019, no. 12, pp. 124018, 2019.

\bibitem{qu2022generalized}
Zhe Qu, Xingyu Li, Rui Duan, Yao Liu, Bo~Tang, and Zhuo Lu,
\newblock ``Generalized federated learning via sharpness aware minimization,''
\newblock in {\em International Conference on Machine Learning}. PMLR, 2022, pp. 18250--18280.

\bibitem{foret2020sharpness}
Pierre Foret, Ariel Kleiner, Hossein Mobahi, and Behnam Neyshabur,
\newblock ``Sharpness-aware minimization for efficiently improving generalization,''
\newblock in {\em International Conference on Learning Representations}, 2020.

\bibitem{buzzega2020dark}
Pietro Buzzega, Matteo Boschini, Angelo Porrello, Davide Abati, and Simone Calderara,
\newblock ``Dark experience for general continual learning: a strong, simple baseline,''
\newblock {\em Advances in neural information processing systems}, vol. 33, pp. 15920--15930, 2020.

\bibitem{parisi2019continual}
German~I Parisi, Ronald Kemker, Jose~L Part, Christopher Kanan, and Stefan Wermter,
\newblock ``Continual lifelong learning with neural networks: A review,''
\newblock {\em Neural Networks}, vol. 113, pp. 54--71, 2019.

\bibitem{li2017learning}
Zhizhong Li and Derek Hoiem,
\newblock ``Learning without forgetting,''
\newblock {\em IEEE transactions on pattern analysis and machine intelligence}, vol. 40, no. 12, pp. 2935--2947, 2017.

\bibitem{mao2021continual}
Fubing Mao, Weiwei Weng, Mahardhika Pratama, and Edward Yapp~Kien Yee,
\newblock ``Continual learning via inter-task synaptic mapping,''
\newblock {\em Knowledge-Based Systems}, vol. 222, pp. 106947, 2021.

\bibitem{sun2023exemplar}
Wenju Sun, Qingyong Li, Jing Zhang, Danyu Wang, Wen Wang, and YangLi-ao Geng,
\newblock ``Exemplar-free class incremental learning via discriminative and comparable parallel one-class classifiers,''
\newblock {\em Pattern Recognition}, vol. 140, pp. 109561, 2023.

\bibitem{yao2010boosting}
Yi~Yao and Gianfranco Doretto,
\newblock ``Boosting for transfer learning with multiple sources,''
\newblock in {\em 2010 IEEE computer society conference on computer vision and pattern recognition}. IEEE, 2010, pp. 1855--1862.

\bibitem{yoon2017lifelong}
Jaehong Yoon, Eunho Yang, Jeongtae Lee, and Sung~Ju Hwang,
\newblock ``Lifelong learning with dynamically expandable networks,''
\newblock in {\em International Conference on Learning Representations}, 2018.

\bibitem{shin2017continual}
Hanul Shin, Jung~Kwon Lee, Jaehong Kim, and Jiwon Kim,
\newblock ``Continual learning with deep generative replay,''
\newblock in {\em Proceedings of the 31st International Conference on Neural Information Processing Systems}, 2017, pp. 2994--3003.

\bibitem{chaudhry2018efficient}
Arslan Chaudhry, Marc’Aurelio Ranzato, Marcus Rohrbach, and Mohamed Elhoseiny,
\newblock ``Efficient lifelong learning with a-{GEM},''
\newblock in {\em International Conference on Learning Representations}, 2019.

\bibitem{aljundi2019online}
Rahaf Aljundi, Eugene Belilovsky, Tinne Tuytelaars, Laurent Charlin, Massimo Caccia, Min Lin, and Lucas Page-Caccia,
\newblock ``Online continual learning with maximal interfered retrieval,''
\newblock in {\em Advances in Neural Information Processing Systems 32}, H.~Wallach, H.~Larochelle, A.~Beygelzimer, F.~d\textquotesingle Alch\'{e}-Buc, E.~Fox, and R.~Garnett, Eds., pp. 11849--11860. Curran Associates, Inc., 2019.

\bibitem{lao2021focl}
Qicheng Lao, Mehrzad Mortazavi, Marzieh Tahaei, Francis Dutil, Thomas Fevens, and Mohammad Havaei,
\newblock ``Focl: Feature-oriented continual learning for generative models,''
\newblock {\em Pattern Recognition}, vol. 120, pp. 108127, 2021.

\bibitem{martins2023meta}
Vinicius~Eiji Martins, Alberto Cano, and Sylvio~Barbon Junior,
\newblock ``Meta-learning for dynamic tuning of active learning on stream classification,''
\newblock {\em Pattern Recognition}, vol. 138, pp. 109359, 2023.

\bibitem{tran2023sharpness}
Lam Tran~Tung, Viet Nguyen~Van, Phi Nguyen~Hoang, and Khoat Than,
\newblock ``Sharpness and gradient aware minimization for memory-based continual learning,''
\newblock in {\em Proceedings of the 12th International Symposium on Information and Communication Technology}, 2023, pp. 189--196.

\bibitem{hinton2015distilling}
Geoffrey Hinton, Oriol Vinyals, and Jeff Dean,
\newblock ``Distilling the knowledge in a neural network,''
\newblock {\em arXiv preprint arXiv:1503.02531}, 2015.

\bibitem{qu2023prevent}
Zhe Qu, Xingyu Li, Xiao Han, Rui Duan, Chengchao Shen, and Lixing Chen,
\newblock ``How to prevent the poor performance clients for personalized federated learning?,''
\newblock in {\em Proceedings of the IEEE/CVF Conference on Computer Vision and Pattern Recognition}, 2023, pp. 12167--12176.

\bibitem{lecun1998gradient}
Yann LeCun, L{\'e}on Bottou, Yoshua Bengio, Patrick Haffner, et~al.,
\newblock ``Gradient-based learning applied to document recognition,''
\newblock {\em Proceedings of the IEEE}, vol. 86, no. 11, pp. 2278--2324, 1998.

\bibitem{krizhevsky2009learning}
Alex Krizhevsky and Geoffrey Hinton,
\newblock ``Learning multiple layers of features from tiny images,''
\newblock Tech. {R}ep.~0, University of Toronto, Toronto, Ontario, 2009.

\bibitem{Le2015TinyIV}
Ya~Le and Xuan~S. Yang,
\newblock ``Tiny imagenet visual recognition challenge,''
\newblock 2015.

\bibitem{boschini2022class}
Matteo Boschini, Lorenzo Bonicelli, Pietro Buzzega, Angelo Porrello, and Simone Calderara,
\newblock ``Class-incremental continual learning into the extended der-verse,''
\newblock {\em IEEE Transactions on Pattern Analysis and Machine Intelligence}, 2022.

\bibitem{schwarz2018progress}
Jonathan Schwarz, Wojciech Czarnecki, Jelena Luketina, Agnieszka Grabska-Barwinska, Yee~Whye Teh, Razvan Pascanu, and Raia Hadsell,
\newblock ``Progress \& compress: A scalable framework for continual learning,''
\newblock in {\em International Conference on Machine Learning}. PMLR, 2018, pp. 4528--4537.

\bibitem{he2016deep}
Kaiming He, Xiangyu Zhang, Shaoqing Ren, and Jian Sun,
\newblock ``Deep residual learning for image recognition,''
\newblock in {\em Proceedings of the IEEE conference on computer vision and pattern recognition}, 2016, pp. 770--778.

\end{thebibliography}

\end{document}